\documentclass[twocolumn]{isrpaper}
\usepackage{adjustbox}

\usepackage{lmodern}

\usepackage{orcidlink}

\usepackage{placeins}

\usepackage{etoolbox}
\AtBeginEnvironment{tabular}{\scriptsize}
\usepackage[
    backend=biber,
    style=ieee,
    natbib=true,
    doi=false,
    url=false,
    isbn=false,
    maxbibnames=99,
    uniquelist=false,
]{biblatex}

\AtEveryBibitem{%
   \clearfield{Title}%
   \clearfield{eprint}%
   \clearfield{note}%
}

\renewcommand{\citet}[1]{\textcite{#1}}
\renewcommand{\citep}[1]{\parencite{#1}}

\title{Walking Speed and Step Length in Clinical Gait from a Moving Handheld Smartphone}

\author{
J.D. Peiffer\,\orcidlink{0000-0003-2382-8065}\,$^{1,2}$, Kunal Shah$^{1}$, Shawana Anarwala$^{1}$, Kayan Abdou$^{1}$, Rujvee Patel$^{3}$, Brenton Pennicooke$^{3}$, R. James Cotton\,\orcidlink{0000-0001-5714-1400}\,$^{1,4}$}

\affiliations{
$^{1}$Shirley Ryan AbilityLab \\ $^{2}$Department of Biomedical Engineering, Northwestern University \\ $^{3}$Department of Neurological Surgery, Washington University School of Medicine in St. Louis \\ $^{4}$Department of Physical Medicine and Rehabilitation, Northwestern University}

\paperdate{November 2026}

\correspondence{rcotton@sralab.org}

\paperabstract{Walking speed is a functional vital sign, yet measuring it requires an instrumented walkway or a timed test on a
marked course. A handheld smartphone easily records walking, but measuring speed from one video requires two difficult
steps: separating the camera's motion from the person's and recovering metric scale. We combine two feed-forward
neural networks, one reconstructing the camera path and scene and one estimating the person's biomechanical pose
relative to the camera, with a sequence solver that refines camera, scale and body over the walk. We tune the solver
on 25 control participants and compare it with four world-frame human-motion methods against motion capture on 48
held-out participants with neurologic conditions or lower-limb prostheses. Without calibration or the participant's
height, our method recovers metric scale to within 2.6\% and walking speed to within 0.016 m/s, outperforming naive
scale correction and low-pass filtering. This error is well below the minimal clinically important differences for
stroke and spinal cord injury (0.15--0.16 m/s) and 69\% below the best comparison method. With the recorded height, it
falls to 0.012 m/s. Our method also measures step length, stride length and step width with biases within about 1 cm
of an instrumented walkway and keeps joint angles within about 6° of motion capture. At two outpatient clinics, its
walking speed tracked same-day clinical scores in patients with cervical myelopathy (Spearman $\rho$ 0.66).}

\begin{document}

\maketitle

\section{Introduction}

\begin{figure*}[t]
\centering
\includegraphics[width=\textwidth]{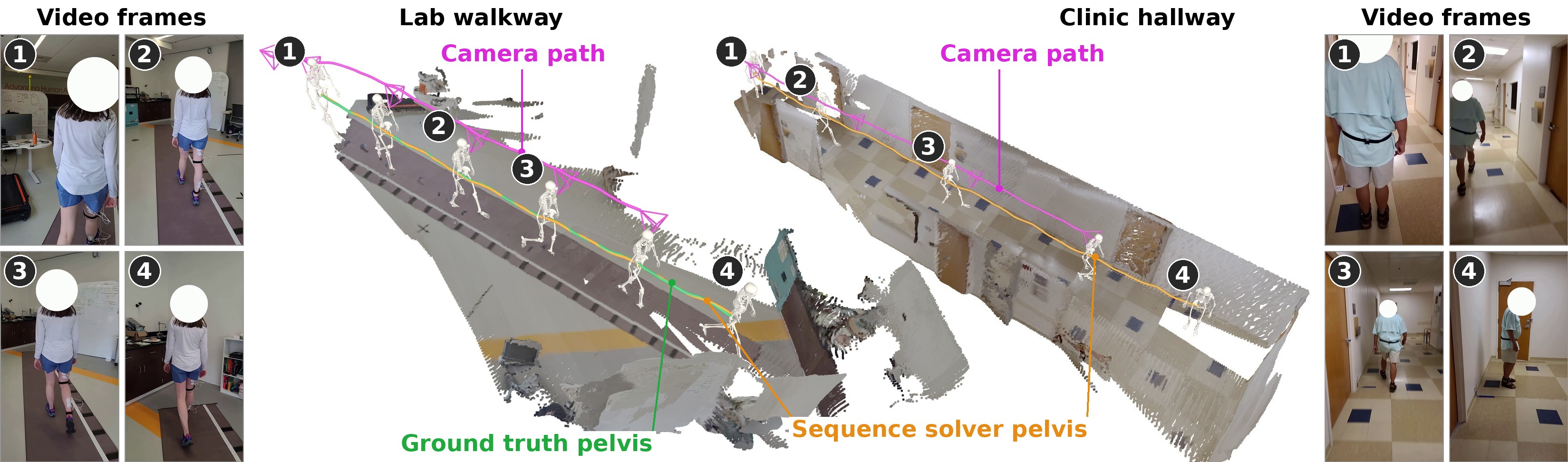}
\vspace{-18pt}
\caption{A walker (pelvis path in orange) and the handheld phone following them (path in magenta), tracked in the world from its video alone. Left: a control walk on the instrumented walkway, with ground truth motion-capture pelvis (green) rigidly aligned. Right: a walk in a clinic hallway with no accompanying ground truth. Outer tiles: video frames, numbered with the skeletons they show.}
\label{fig-world-render}
\vspace{-4pt}
\end{figure*}
Walking speed is part of the core set of
outcome measures in rehabilitation \parencite{moore_core_2018} and has been proposed as the sixth vital sign \parencite{fritz_walking_2009}. It predicts survival in older adults
\parencite{studenski_gait_2011} and tracks recovery after neurological injury, with minimal clinically important
differences (MCIDs) of 0.16 and 0.15 m/s after stroke and incomplete spinal cord injury
\parencite{tilson_meaningful_2010, forrest_are_2014}, and smaller changes of 0.04--0.06 m/s still considered meaningful
in older adults and after stroke \parencite{perera_meaningful_2006}.
Yet it is generally measured only under controlled conditions, by a clinician timing a marked course with a
stopwatch or on an instrumented walkway \parencite{bilney_concurrent_2003}.

Handheld smartphone video could measure walking speed anywhere someone can walk, but existing smartphone-based
biomechanics approaches either fix the camera on a tripod \parencite{uhlrich2023opencap, gilon_opencap_2026} or track
movement only relative to the camera \parencite{peiffer_portable_2026}. A fixed tripod requires room to set up and sees a
person well only over a short pass before they shrink in the frame or leave the field of view. Walking behind
someone with a handheld smartphone keeps them well in view at a consistent image resolution over longer walks, including
down clinic hallways. However, tracking a moving person in the world from a moving camera is a difficult problem: a
person's displacement in the image mixes their own motion with the camera's, and monocular video carries no metric scale.
World-frame human mesh-estimation methods \parencite{shin_wham_2024, shen_gvhmr_2024, chen_human3r_2026} recover a person
trajectory from moving-camera video, but they are scored mainly on short-horizon or displacement-normalized metrics in
populations without gait impairments; none report metric walking speed, and none have been tested within a clinical
workflow. Feed-forward geometry models \parencite{wang_vggt_2025, wang_vggt_omega_2026} track the camera and scene depth in
one pass, up to an unknown scale, but not the person.

Building on the Portable Biomechanics Laboratory's handheld workflow \parencite{peiffer_portable_2026}, we also
recover the camera path and metric scale by composing state-of-the-art foundation models with a world-space sequence
solver (Figure~\ref{fig-world-render}). First, a visual geometry model, VGGT-$\Omega$
\parencite{wang_vggt_omega_2026}, reconstructs the unscaled camera path and per-frame depth maps. Next, Biomechanical 3D
Body (B3D) \parencite{cotton2026biomechanical} estimates the body in each frame's camera coordinates. Finally, a single
scale factor between VGGT-$\Omega$'s depth maps and B3D's body meshes puts the scene and camera path in meters, and our
world-space sequence solver refines the camera path, scale and body jointly over the whole walk.

In summary, our contributions are:

\begin{itemize}
\item A calibration-free workflow for handheld smartphone gait video that measures full body kinematics and walking speed on held-out clinical participants to 0.016 m/s, far below published MCIDs, and to 0.012 m/s given their recorded height.
\item A benchmark against four world-frame human-motion methods (GVHMR \parencite{shen_gvhmr_2024}, FlowHMR \parencite{wang_flowhmr_2026}, Human3R \parencite{chen_human3r_2026} and JOSH \parencite{liu_josh_2026}), which our workflow outperforms: 69\% lower walking-speed error than the best method and 54\% lower world-frame joint error.
\item A feasibility test in routine clinical care: in two neurosurgery clinics, walking speed from our method tracks the clinical scores of patients with cervical myelopathy.
\end{itemize}

\section{Related work}

\textbf{Deployable biomechanical measurement.} OpenCap \parencite{uhlrich2023opencap} estimates movement kinematics and
dynamics from two calibrated smartphones, and OpenCap Monocular \parencite{gilon_opencap_2026} from one on a tripod,
refining WHAM's \parencite{shin_wham_2024} world-frame estimate into a biomechanical model.
The Portable Biomechanics Laboratory (PBL) \parencite{peiffer_portable_2026} estimates joint angles from a moving handheld
smartphone, making it more deployable, but recovers only the
body's global rotation from the phone's orientation sensors, missing its global translation. Biomechanical 3D Body (B3D),
which extends SAM 3D Body \parencite{yang_sam_2026} with a biomechanical head, rapidly estimates biomechanics from each
frame of monocular video in a feed-forward manner. In a head-to-head
comparison, both PBL and B3D estimated joint angles more accurately than OpenCap Monocular \parencite{cotton2026biomechanical}.

\textbf{World-frame human motion.} From a single RGB video, GVHMR \parencite{shen_gvhmr_2024} predicts global human meshes in a
gravity-aligned frame, and FlowHMR \parencite{wang_flowhmr_2026} generates them with a video-conditioned flow-matching
model post-trained to favor physically trackable motion; neither outputs a camera path. JOSH \parencite{liu_josh_2026}
optimizes the camera, the scene and the human mesh in a single stage, tied together by human-scene contact, and
Human3R \parencite{chen_human3r_2026} predicts the camera, the scene and every person's mesh jointly.

\textbf{Feed-forward geometry.} VGGT \parencite{wang_vggt_2025} regresses cameras, depth and pointmaps for many views in
one pass, and VGGT-$\Omega$ \parencite{wang_vggt_omega_2026} improves its efficiency and accuracy.

\section{Methods}\label{methods}

From a handheld smartphone video of a person walking, we estimate their pelvis trajectory and body pose in a metric
world frame in two stages. First, we build a per-frame world-space estimate: VGGT-$\Omega$ reconstructs the camera
path, B3D estimates the person's biomechanical pose and body mesh relative to the camera, and that mesh sets the
reconstruction's metric scale. Second, a sequence solver jointly optimizes the camera path, scale and body over the
whole walk.

\subsection{Per-Frame World-Space Estimate}

\textbf{Camera path and depth.} A feed-forward geometry model, VGGT-$\Omega$ \parencite{wang_vggt_omega_2026}, takes every frame of the walking
bout in a single forward pass and predicts per-frame camera poses, fields of view and dense depth maps,
consistent with one another but at an arbitrary non-metric scale. One VGGT-$\Omega$
pass fits up to about 440 frames on a 48 GB GPU (at its 688$\times$384 input), so longer walks run in 100-frame
windows overlapping by 50 frames, each aligned to the camera path built so far by a similarity transform fitted on
their shared frames. Instead of calibrating the phone's camera, we convert
VGGT-$\Omega$'s fields of view into one focal length $f$ per walk (the mean of the per-axis median focal lengths,
principal point at the image center), used in all later stages.

\textbf{Person estimate.} On the subject's bounding box tracked by PosePipe \parencite{cotton_posepipe_2022}, B3D
\parencite{cotton2026biomechanical} estimates the camera-relative person feed-forward in each frame, outputting a metric
Momentum Human Rig (MHR) \parencite{ferguson_mhr_2025} body mesh, a biomechanical skeleton with segment scales and joint
angles, and 3D anatomical surface markers. Its per-frame body segment scales are then replaced by
their median over the bout, so one consistent body is posed in every frame. We run B3D with the estimated focal length $f$ as its camera intrinsics, which
condition the network through its camera-ray embedding and crop token. B3D's camera head regresses a weak-perspective crop scale
$s_{\text{cam}}$, which sets the person's metric depth in the camera frame as $d = 2f / (b\,s_{\text{cam}})$ for
bounding-box size $b$. The resulting root translation places both the MHR mesh and the biomechanical skeleton.

\textbf{Metric scale.} We recover an initial metric scale $s_0$ of the scene by comparing VGGT-$\Omega$'s arbitrary-scale scene
depth map with the metric depth of B3D's MHR mesh on the pixels where the person is visible. To do
so, we render the MHR mesh into each frame at focal length $f$ and pair its depth $z^{\text{M}}_p$ with
VGGT-$\Omega$'s scene depth $z^{\text{V}}_p$ on the pixels $p$ inside its silhouette, eroded by 3 px to avoid depth
boundaries. In each frame $t$, a median-absolute-deviation gate (3 MAD) on their ratio discards outliers,
and a least-squares fit on the remaining pixels $\mathcal{I}_t$ gives that frame's scale,
$s_t = \arg\min_s \sum_{p \in \mathcal{I}_t} \bigl(z^{\text{M}}_p - s\,z^{\text{V}}_p\bigr)^2$, and
$s_0 = \operatorname{median}_t\, s_t$.

\subsection{Sequence Solver}

The per-frame estimate has errors due to B3D's frame-to-frame pose and depth jitter, VGGT-$\Omega$'s noisy
camera path and the mesh-depth error in its scale $s_0$. To reduce these
errors, we jointly optimize the person's pose and world trajectory, the camera path and the metric scale over the
whole walk, starting from the per-frame estimate. We call this optimization the sequence solver.

\textbf{Unknowns.} The solver explicitly parameterizes frame-by-frame corrections to B3D's pelvis translation, pelvis
rotation and joint angles ($\delta\mathbf{q}_t$) and to the camera's
rotation and position ($\delta\boldsymbol{\omega}_t$, $\delta\mathbf{c}_t$).
B3D's body segment scales stay fixed, and one metric scale $s$ of the VGGT-$\Omega$ reconstruction is optimized over
the whole walk. The camera-relative body is metric already, so $s$ scales only the camera path: the
camera's metric position is $\mathbf{c}_t = s\,\mathbf{C}_t$, with $\mathbf{C}_t$ its corrected
position. A body point $\mathbf{P}_t$ in the camera frame is then placed in the world as
$\mathbf{X}_t = \mathbf{c}_t + \mathbf{R}_t \mathbf{P}_t$, with $\mathbf{R}_t$ the corrected camera rotation.

\textbf{Energy.} Each term divides its residuals by a standard deviation $\sigma$ and penalizes them quadratically or
with the Huber penalty $\rho(\cdot)$, linear beyond 2--3$\sigma$. Accelerations and
velocities are second and central differences over consecutive frames. The $\sigma$ were tuned on the control walks and applied unchanged to the
clinical walks.

\textit{B3D's observations.} For each frame $t$ and marker $k$, we compare the skeleton's mean-centered 3D marker position
$\mathbf{y}_{tk}$ with B3D's own 3D marker target $\mathbf{y}^{*}_{tk}$ ($\sigma_M$ = 2 cm), and its projected 2D
position $\mathbf{u}_{tk}$ with the target's $\mathbf{u}^{*}_{tk}$ ($\sigma_u$ = 8.9 px, B3D's measured image
residual),

\begin{equation}
\begin{aligned}
E_{\text{obs}} = \tfrac{1}{2} \sum_{t,k} \bigl[ & \rho(\lVert \mathbf{y}_{tk} - \mathbf{y}^{*}_{tk} \rVert / \sigma_M) \\
& + \rho(\lVert \mathbf{u}_{tk} - \mathbf{u}^{*}_{tk} \rVert / \sigma_u) \bigr] .
\end{aligned}
\end{equation}

\textit{Joint angles.} The corrected joint angles $\mathbf{q}_t$ stay smooth and their corrections $\delta\mathbf{q}_t$ small
($\sigma_{\ddot{q}}$ = 60 rad/s$^2$, $\sigma_q$ = 0.3 rad),

\begin{equation}
E_q = \sum_t \frac{\lVert \ddot{\mathbf{q}}_t \rVert^2}{\sigma_{\ddot{q}}^2} +
\frac{\lVert \delta\mathbf{q}_t \rVert^2}{\sigma_q^2} .
\end{equation}

\textit{Camera.} For each frame $t$, we keep the camera's rotation and position corrections $\delta\boldsymbol{\omega}_t$
and $\delta\mathbf{c}_t$ small ($\sigma_\omega$ = 0.05°, $\sigma_c$ = 0.10 m), and its motion smooth through
the acceleration of its metric position $\mathbf{c}_t$ ($\sigma_{\ddot{c}}$ = 3 m/s$^2$) and its
angular acceleration $\boldsymbol{\alpha}_t$ ($\sigma_\alpha$ = 3 rad/s$^2$),

\begin{equation}
E_{\text{cam}} = \sum_t \frac{\lVert \delta\boldsymbol{\omega}_t \rVert^2}{\sigma_\omega^2} +
\frac{\lVert \delta\mathbf{c}_t \rVert^2}{\sigma_c^2} +
\frac{\lVert \ddot{\mathbf{c}}_t \rVert^2}{\sigma_{\ddot{c}}^2} +
\frac{\lVert \boldsymbol{\alpha}_t \rVert^2}{\sigma_\alpha^2} .
\end{equation}

\textit{Pelvis.} We penalize the world acceleration of the pelvis $\mathbf{p}_t$ ($\sigma_{\ddot{p}}$ = 3 m/s$^2$),

\begin{equation}
E_{\text{pel}} = \sum_t \rho(\lVert \ddot{\mathbf{p}}_t \rVert / \sigma_{\ddot{p}}) .
\end{equation}

\textit{Planted feet.} We use the Gait Transformer \parencite{cotton_transforming_2022} to detect each foot's stance frames
$\mathcal{S}$ from B3D's 3D joints. In these frames, a planted-foot term penalizes the world motion of the contact point
$\mathbf{x}_{tj}$ of foot $j$, defined as the slower moving of its heel and toe ($\sigma_v$ = 0.2 m/s), and a floor term
penalizes the deviation of its height $z_{tj}$ above the floor plane fitted to that frame's VGGT-$\Omega$ points from $h_0$ = 1.5 cm
($\sigma_n$ = 5 cm),

\begin{equation}
E_{\text{foot}} = \sum_{(t,j) \in \mathcal{S}} \bigl[\rho(\lVert \dot{\mathbf{x}}_{tj} \rVert / \sigma_v) +
\rho\bigl((z_{tj} - h_0) / \sigma_n\bigr)\bigr] .
\end{equation}

\textit{Scale.} A log-normal prior holds $s$ near $s_0$ ($\sigma_s$ = 15\%),

\begin{equation}
E_s = \bigl(\log(s / s_0) / \sigma_s\bigr)^2 .
\end{equation}

\textbf{Recorded height.} Clinics record height routinely, so the solver can optionally constrain the detected person
to the participant's height. The body is rescaled uniformly by a factor $h$, placing it as
$\mathbf{X}_t = \mathbf{c}_t + h\,\mathbf{R}_t \mathbf{P}_t$, with the term
$E_h = (\log(h\,\ell / H) / \sigma_h)^2$ ($\sigma_h$ = 2\%), which holds the skeleton's standing height $\ell$
to the recorded height $H$.

\textbf{Optimization.} We minimize the sum of the energy terms over all unknowns jointly with L-BFGS
\parencite{liu_limited_1989} for 100 iterations (400 with a recorded height, which adds $E_h$ and whose rescaling needs
them to converge); more iterations had a negligible effect on every reported metric. The solver is implemented in JAX
\parencite{jax_2018}, and each walk has its own parameters, with several solved in parallel for speed. A walk whose
initial energy per frame exceeds twice the largest on the control walks is refused and counted as a failure.

\subsection{Runtime}

On one NVIDIA B200 GPU, the workflow and solver process a walk, given the subject's tracked box, at about
0.3 s per video frame, 8--9 times the
walk's duration: 54 s for a 7 s walk (210 frames) and 3.2 min for a 22 s walk (658 frames). VGGT-$\Omega$ and B3D
take about three quarters of this time; the solve takes about 3 s per sequence, and the mesh-depth scale, floor fits
and foot contacts the rest. Loading the models and compiling
add about 9 min on first use. On the same GPU and walks, the comparison methods take 0.03 s per frame (GVHMR), 0.2 s
(Human3R), 0.44 s (FlowHMR, including its VGGT-$\Omega$ camera estimate) and 1.6 s (JOSH).

\subsection{Comparison Methods}

We run GVHMR \parencite{shen_gvhmr_2024}, FlowHMR \parencite{wang_flowhmr_2026}, Human3R \parencite{chen_human3r_2026} and
JOSH \parencite{liu_josh_2026} on the same walks for comparison, each from its released code. GVHMR and FlowHMR receive VGGT-$\Omega$'s focal length; GVHMR's
default focal length, the image diagonal, is 54\% too long. FlowHMR also receives the subject's box and
VGGT-$\Omega$'s camera rotations. Human3R runs in its long-sequence configuration, reset every 100 frames. JOSH runs
in its 100-frame chunks, optimizing a keyframe every 5 frames. These four predict the person in their own world
frame. Human3R's and JOSH's cameras are scored at their native scale; GVHMR and FlowHMR output no camera path.
For multi-person methods, we keep the person best overlapping the tracked box. We also compare joint
angles and positions with PBL \parencite{peiffer_portable_2026} on the same videos.

We also compare the solver with two naive corrections of the per-frame estimate's scale bias and jitter, alone and
combined: one scale constant (0.935, fitted on the control walks) and a zero-phase low-pass filter (3 Hz cutoff, chosen
on the control walks). Finally, we ablate energy terms from the height-free solver.

\subsection{Datasets and Evaluation}

\textbf{Lab recordings.} We use 25 healthy controls, 28 people with neurologic conditions
and 25 lower-limb prosthesis users, from the in-lab Portable
Biomechanics Laboratory study \parencite{peiffer_portable_2026}, walking at a self-selected speed in our hospital
laboratory with their usual assistive devices. A researcher followed each participant from behind with a handheld
smartphone (Samsung Galaxy S8 or S20 Ultra, 30 Hz, 1080$\times$1920). We use one walk per participant. The study was approved by the Northwestern
University Institutional Review Board, with informed consent from every participant.

\textbf{Clinic recordings.} To evaluate feasibility in a clinical setting, we also recorded patients with degenerative
cervical myelopathy at one or more visits to outpatient neurosurgery clinics \parencite{peiffer_portable_2026}. After
their physician consultation, they walked in one of two clinic hallways, which differ in lighting, room geometry and
texture, while a researcher followed them with the handheld phone. Each session contributes one self-selected walking
video (8--30 s, median 15 s) and is paired with the same-day modified Japanese Orthopaedic Association (mJOA) score
\parencite{benzel_cervical_1991} and its lower-extremity motor subscore (0--7), an ordinal scale. This study was approved
by the Northwestern University and Washington University in St. Louis Institutional Review Boards.

\textbf{Ground truth.} For the lab recordings only, ground-truth biomechanical pose and pelvis trajectory come from a markerless multi-camera recording \parencite{cotton_optimizing_2023} and reconstruction
\parencite{cotton_differentiable_2025} system at 29 Hz, whose biomechanical model shares B3D's leg joint angles. For camera tracking accuracy, we segment the
smartphone in the multi-camera recordings with SAM 3 \parencite{carion_sam3_2025} and triangulate its trajectory,
dropping and interpolating samples with over 100 px reprojection error or implied speeds over 3 m/s.
A GAITRite instrumented walkway \parencite{bilney_concurrent_2003} provides the ground truth for step length, stride
length and step width.

\textbf{Metrics.} We report per-walk medians over whole walking bouts. The solver
failed on two clinical walks, and three more were dropped because a comparison method's output was incomplete
(JOSH on two, Human3R on one), so the clinical results cover 48 walks: 25 participants with neurologic conditions
(14 with an assistive device) and 23 prosthesis users (2 with one). The metrics fall in six groups (Table~\ref{tab-phone-gait}); the first three are at each method's own scale:

\begin{itemize}
\item \textit{Scale} (\%): pelvis-path scale error (its size error relative to ground truth after similarity alignment) and the
camera path's, likewise.
\item \textit{Walking speed} (m/s): error in the walking speed over the whole bout, and root-mean-square (RMS) speed error over 0.25 s
and 1 s windows.
\item \textit{Trajectory}: pelvis relative trajectory error (RTE, \% of path length after rigid alignment), jitter ratio (pelvis
RMS acceleration over the ground truth's; 1 is ideal) and camera absolute trajectory error (ATE, mm).
\item \textit{Similarity-aligned} (mm, RTE in \%), after fitting rotation, translation and scale to ground truth over the bout: RTE (drift) and the
world-aligned mean per-joint position error (WA-MPJPE); and the world MPJPE (W-MPJPE), with the same alignment
fitted on the first two frames only, so that it accumulates drift.
\item \textit{Per-frame aligned} (mm, joint angles in degrees): Procrustes-aligned MPJPE (PA-MPJPE) on 12 limb joints (shoulders, elbows, wrists, hips, knees,
ankles); and, for B3D and PBL, whose leg joint angles the ground truth shares, the mean absolute error (MAE) of hip,
knee and ankle angles.
\item \textit{Foot contact} (mm): ankle slide per flat-foot phase (forefoot contact to heel-off, timed by the walkway, independent of
the solver's contacts) and the spread of the planted toe's height across footfalls, each against the ground truth's
own value.
\end{itemize}

\textbf{Spatial gait parameters.} Step length, stride length and step width (cm) are computed from the solver's heel
position, with the recorded height, at each footfall detected by the Gait Transformer, using the walkway's definitions (step width as the heel-to-heel base of support). Each walk's mean is compared with
the walkway's as bias (estimate $-$ walkway), 95\% limits of agreement (LoA), intraclass correlation (ICC(2,1)) and mean absolute
error (MAE).

\textbf{Clinic walking speed.} We process each clinic video with the participant's recorded height.
Without a walkway marking a test segment, we use each video's steady walking speed: the median speed of the solver's
pelvis over 2 s windows whose net displacement is at least 0.9 of their path length, between the first and last foot
contact. Of the 89 sessions with a same-day subscore, the solver failed on 4, leaving 85 sessions of 48
patients, over which we report Spearman's $\rho$ between steady speed and each score.

\section{Results}

\subsection{Evaluation on Held-Out Clinical Population}

\begin{table*}
\centering
\caption[]{Gait tracking on 48 clinical walks: per-walk medians, best in bold, second underlined.

\begin{adjustbox}{max width=\linewidth}\begin{tabular}{lrrrrrrr}
\toprule
Metric & Per-frame estimate & Ours & Ours, recorded height & GVHMR & FlowHMR & JOSH & Human3R \\
\midrule
\multicolumn{8}{l}{\textbf{Scale}} \\
\hspace{1em}Pelvis-path scale error (\%) $\downarrow$ & 5.8 & \underline{2.6} & \textbf{2.1} & 11.0 & 7.6 & 8.3 & 17.2 \\
\hspace{1em}Camera-path scale error (\%) $\downarrow$ & 6.0 & \underline{3.2} & \textbf{2.3} & --- & --- & 9.1 & 19.0 \\
\midrule
\multicolumn{8}{l}{\textbf{Walking speed}} \\
\hspace{1em}Speed error (m/s) $\downarrow$ & 0.086 & \underline{0.016} & \textbf{0.012} & 0.057 & 0.051 & 0.084 & 0.085 \\
\hspace{1em}Speed RMSE, 0.25 s windows (m/s) $\downarrow$ & 0.154 & \textbf{0.094} & \underline{0.098} & 0.131 & 0.102 & 0.183 & 0.268 \\
\hspace{1em}Speed RMSE, 1 s windows (m/s) $\downarrow$ & 0.064 & \underline{0.039} & \textbf{0.037} & 0.090 & 0.071 & 0.090 & 0.154 \\
\midrule
\multicolumn{8}{l}{\textbf{Trajectory}} \\
\hspace{1em}RTE (\%) $\downarrow$ & 1.43 & \underline{0.75} & \textbf{0.71} & 3.05 & 2.33 & 2.63 & 4.78 \\
\hspace{1em}Accel. jitter ratio & 6.0 & 0.7 & 0.7 & 0.6 & 0.5 & 2.4 & 7.6 \\
\hspace{1em}Camera ATE (mm; oracle scale 29) $\downarrow$ & 92 & \underline{64} & \textbf{49} & --- & --- & 136 & 349 \\
\midrule
\multicolumn{8}{l}{\textbf{Similarity-aligned}} \\
\hspace{1em}RTE, Sim(3) (\%) $\downarrow$ & 0.40 & \underline{0.36} & \textbf{0.34} & 1.00 & 0.90 & 0.69 & 1.60 \\
\hspace{1em}WA-MPJPE (mm) $\downarrow$ & 67.9 & \underline{60.5} & \textbf{60.3} & 118.5 & 104.5 & 97.4 & 180.5 \\
\hspace{1em}W-MPJPE (mm) $\downarrow$ & 385.3 & \underline{299.4} & \textbf{286.9} & 886.2 & 899.9 & 656.6 & 795.0 \\
\midrule
\multicolumn{8}{l}{\textbf{Per-frame aligned (pose only)}} \\
\hspace{1em}PA-MPJPE (mm) $\downarrow$ & 46.6 & \textbf{45.8} & \textbf{45.8} & 49.6 & 49.5 & 53.1 & 54.8 \\
\hspace{1em}Joint-angle MAE (deg) $\downarrow$ & 5.7 & \textbf{5.6} & \textbf{5.6} & --- & --- & --- & --- \\
\midrule
\multicolumn{8}{l}{\textbf{Foot contact}} \\
\hspace{1em}Ankle slide (mm; ground truth 26) & 51 & 32 & 28 & 20 & 28 & 80 & 106 \\
\hspace{1em}Footfall height spread (mm; ground truth 4) & 10 & 10 & 11 & 12 & 6 & 19 & 27 \\
\bottomrule
\end{tabular}\end{adjustbox}}
\label{tab-phone-gait}
\end{table*}

A world-frame movement trajectory can be wrong in three ways: scale (the whole trajectory too large or too small),
jitter (frame-to-frame noise) and drift (incorrect trajectory shape ignoring scale). On the 48 held-out clinical
participants (Table~\ref{tab-phone-gait}), our method has the lowest error of any method in scale, walking speed, drift as well as joint positions.

\textbf{Scale.} Our initial per-frame estimate has the lowest pelvis-path scale error of any method (5.8\%, against 7.6--17.2\%),
and the solver more than halves it even without the participant's height (2.6\%; 2.1\% with the recorded height).
The scene-based JOSH and Human3R lose mostly on scale (8.3 and 17.2\%). The solver barely changes the camera path's
shape (29 mm ATE at ground-truth scale before, 27 mm after), so its gain in camera ATE (92 to 64 mm, and 49 mm with
the height) comes from the scale.

\begin{figure}[!t]
\centering
\includegraphics[width=\columnwidth]{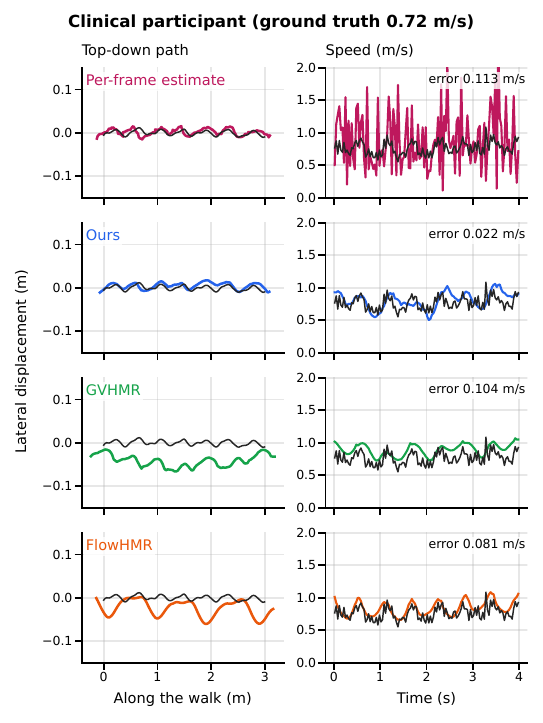}
\vspace{-23pt}
\caption{Pelvis path seen from above and speed over the middle 4 s of one clinical participant's walk. One row per method, motion-capture ground truth as a thin black line. Each method is rigidly aligned to ground truth over the whole walk (rotation and translation, no scale), so it keeps its own metric scale. Each speed panel gives the whole walk's walking-speed error.}
\label{fig-person-trajectory}
\end{figure}

\textbf{Walking speed.} The solver lowers the per-frame estimate's walking-speed error from 0.086 to 0.016 m/s
(0.012 m/s with the recorded height), 69\% below the best comparison method, FlowHMR (0.051 m/s; GVHMR 0.057, JOSH
0.084, Human3R 0.085 m/s). The per-frame estimate's error comes mostly from jitter: its pelvis accelerates six times more
than the ground truth's, which puts it above the smooth temporal models and buries the gait cycle
(Figure~\ref{fig-person-trajectory}). The solver removes this jitter and recovers the gait cycle, and within the walk
our method also has the lowest speed error over windows of about a step (0.25 s) and a stride (1 s).

Across speeds (Figure~\ref{fig-clinical-speed}a), the error with the recorded height is proportional to speed (median
+0.8\%), so it is smallest for the slowest walkers, where small changes matter most: 0.009 m/s median for the 28
walking below 0.8 m/s, and 0.008 m/s for the 16 who walked with an assistive device (median speed 0.37 m/s). By
group, the error is 0.009 m/s for the participants with neurologic conditions and 0.018 m/s for the prosthesis users,
who walk faster (median 0.53 and 0.81 m/s). The bias is +0.010 m/s (95\% limits of agreement $-$0.04 to +0.06 m/s,
ICC 0.997), and the 25 control walks match it (+0.02 m/s; $-$0.03 to +0.07 m/s, ICC 0.996).

\textbf{Trajectory and drift.} The per-frame estimate drifts less than any comparison method (RTE 1.43\%), and the solver reduces
this to 0.75\% (0.71\% with the recorded height). GVHMR and FlowHMR are smooth (jitter ratio 0.5--0.6), but their
scale is off and their drift accumulates, so their world-frame joint error grows to 886--900 mm over a walk. Once a
similarity alignment removes scale, JOSH drifts less than GVHMR and FlowHMR (RTE 0.69 against 0.90--1.00\%). World-space
joint error depends mostly on the trajectory, and ours is the lowest of any method (WA-MPJPE 60.3 mm, W-MPJPE 287 mm,
against 385--900 mm).

\textbf{Joint angles and positions.} The solver changes B3D's joint angles and positions very little: our hip, knee and
ankle angles are within 5.6° MAE of the motion-capture biomechanical fit, against 5.7° for the
per-frame estimate, and, after aligning each frame's skeleton to the ground truth, our joint positions are within
45.8 mm (PA-MPJPE), against 46.6 mm. Our joint positions are more accurate than the SMPL-based methods'
(49.5--54.8 mm), but on the same walks PBL, which uses the phone's calibrated intrinsics and recovers no metric scale,
is more accurate in both joint angles (4.9° MAE) and joint positions (PA-MPJPE 37.6 mm). The world-frame
comparison methods predict meshes and give no comparable joint angles.

\textbf{Foot contact.} The solver brings foot slide close to the ground truth's measurement: while the foot is flat
(forefoot contact to heel-off), its ankle moves 28 mm (32 mm without the height), against 26 mm for the multi-camera
ground truth's ankle, which also moves slightly during this part of the gait cycle. Before the
solver the ankle slides 51 mm, and the scene-based methods slide most (JOSH 80 mm, Human3R 106 mm). GVHMR's ankle
moves least (20 mm), less than the ground truth's, potentially indicating an over-smoothed trajectory (jitter ratio 0.6).
Footfalls stay level to within 10--11 mm for our method and 6 mm for FlowHMR (ground truth 4 mm, Human3R 27 mm).

\textbf{Step length, stride length and step width.} All three spatial parameters agree closely with the walkway on both
control and clinical walks (Table~\ref{tab-spatial-gait}), with biases of +0.6 and +0.2 cm for step length (control and
clinical), +1.2 and +0.2 cm for stride length and 0.0 and $-$0.8 cm for step width. Mean absolute errors are
1.1--1.3 cm for step length and width and 2.0--2.5 cm for stride length. As expected, the clinical participants took
shorter, wider steps than the controls on the walkway (step length 52.6 against 65.0 cm, step width 15.8 against 10.4 cm). One clinical walk was excluded because gait-event segmentation
failed on its very slow, shuffling steps.

\begin{table}
\centering
\caption[]{Spatial gait parameters against the walkway, per walk (cm).

\begin{adjustbox}{max width=\linewidth}\begin{tabular}{llrrr}
\toprule
Measure & Walks & Bias [95\% LoA] & ICC & MAE \\
\hline
Step length & Control (25) & +0.6 [-2.2, +3.4] & 0.995 & 1.1 \\
 & Clinical (47) & +0.2 [-3.1, +3.5] & 0.993 & 1.3 \\
Stride length & Control (25) & +1.2 [-4.1, +6.4] & 0.995 & 2.0 \\
 & Clinical (47) & +0.2 [-6.1, +6.5] & 0.994 & 2.5 \\
Step width & Control (25) & 0.0 [-3.0, +3.0] & 0.900 & 1.1 \\
 & Clinical (47) & -0.8 [-3.8, +2.1] & 0.934 & 1.3 \\
\bottomrule
\end{tabular}\end{adjustbox}}
\label{tab-spatial-gait}
\end{table}

\subsection{Comparison with Naive Corrections and Solver Ablations}

\begin{table*}
\centering
\caption[]{Sequence solver, baseline, naive corrections and solver ablations: per-walk medians, best in bold, second underlined.

\vspace{3pt}\begin{adjustbox}{max width=\linewidth}\begin{tabular}{lrrrrrr|rrr}
\toprule
 & \multicolumn{2}{c}{Sequence solver} & \multicolumn{1}{c}{Baseline} & \multicolumn{3}{c}{Naive corrections} & \multicolumn{3}{c}{No-height ablations} \\ \cmidrule(lr){2-3}\cmidrule(lr){4-4}\cmidrule(lr){5-7}\cmidrule(lr){8-10}
Metric & No height & Recorded height & Per-frame & + constant & + filter & + both & - floor & - pelvis accel. & - foot \\
\midrule
\multicolumn{10}{l}{\textbf{Scale}} \\
\hspace{1em}Pelvis-path scale error (\%) $\downarrow$ & \underline{2.6} & \textbf{2.1} & 5.8 & 3.7 & 5.8 & 3.7 & 9.0 & 2.7 & 2.3 \\
\hspace{1em}Camera-path scale error (\%) $\downarrow$ & \underline{3.2} & \textbf{2.3} & 6.0 & 3.5 & 6.0 & 3.5 & 9.5 & 3.3 & 3.1 \\
\midrule
\multicolumn{10}{l}{\textbf{Walking speed}} \\
\hspace{1em}Speed error (m/s) $\downarrow$ & \underline{0.016} & \textbf{0.012} & 0.086 & 0.049 & 0.031 & 0.020 & 0.043 & 0.022 & 0.012 \\
\hspace{1em}Speed RMSE, 0.25 s windows (m/s) $\downarrow$ & \textbf{0.094} & \underline{0.098} & 0.154 & 0.134 & 0.131 & 0.114 & 0.108 & 0.110 & 0.090 \\
\hspace{1em}Speed RMSE, 1 s windows (m/s) $\downarrow$ & \underline{0.039} & \textbf{0.037} & 0.064 & 0.046 & 0.062 & 0.045 & 0.062 & 0.037 & 0.039 \\
\midrule
\multicolumn{10}{l}{\textbf{Trajectory}} \\
\hspace{1em}RTE (\%) $\downarrow$ & \underline{0.75} & \textbf{0.71} & 1.43 & 1.04 & 1.42 & 1.03 & 2.25 & 0.79 & 0.71 \\
\hspace{1em}Accel. jitter ratio & 0.7 & 0.7 & 6.0 & 5.6 & 0.6 & 0.5 & 0.7 & 2.7 & 0.7 \\
\hspace{1em}Camera ATE (mm) $\downarrow$ & 64 & \textbf{49} & 92 & \underline{63} & 92 & \underline{63} & 148 & 62 & 63 \\
\midrule
\multicolumn{10}{l}{\textbf{Foot contact}} \\
\hspace{1em}Ankle slide (mm; ground truth 26) & 32 & 28 & 51 & --- & --- & --- & 31 & 27 & 35 \\
\bottomrule
\end{tabular}\end{adjustbox}}
\label{tab-phone-variants}
\end{table*}

The solver outperforms naively scaling or low-pass filtering the trajectory, both separately and combined
(Table~\ref{tab-phone-variants}, ``+ constant'', ``+ filter'' and ``+ both''). The ablations in Table~\ref{tab-phone-variants}'s last three columns, each removing one energy term,
show that the floor term sets the metric scale, the pelvis-acceleration term the smoothness, and the planted-foot term
mainly limits foot slide.

\textbf{Scale constant and filtering.} The scale constant fitted on the control walks carries over to new participants,
lowering pelvis-path scale error from 5.8 to 3.7\% and walking-speed error from 0.086 to 0.049 m/s. The filter
removes the jitter (jitter ratio 6.0 to 0.6) but not the scale error, lowering walking-speed error to 0.031 m/s.
Together they reach 0.020 m/s, still above the height-free solver's 0.016 m/s. The solver also leads the combination on
scale and drift (pelvis-path scale error 2.6 against 3.7\%, RTE 0.75 against 1.03\%), though its camera path is no better (64 against 63 mm ATE).

\textbf{Solver Energy Ablations.} Without the
floor term the scale error grows past even the per-frame estimate's (9.0 against 5.8\%), and walking-speed error and
drift rise with it. Without the pelvis-acceleration term the jitter returns (jitter ratio 2.7 against 0.7). On these
walks, removing the planted-foot term slightly improves walking speed and trajectory (walking-speed error 0.012
against 0.016 m/s), at the cost of more foot slide (35 against 32 mm).

\subsection{Walking Speed in the Clinic}

In the clinical cohort recorded in clinic hallways, walking speed from smartphone video tracks clinical
severity. Steady walking speed in the clinic hallway trends
with the same-day mJOA lower-extremity subscore (Figure~\ref{fig-clinical-speed}b; Spearman $\rho$ = 0.66). It tracks
the total mJOA less closely ($\rho$ = 0.57), likely because the combined score also covers upper-limb, sensory and
bladder function.

\section{Discussion}

\begin{figure}[!t]
\centering
\includegraphics[width=\linewidth]{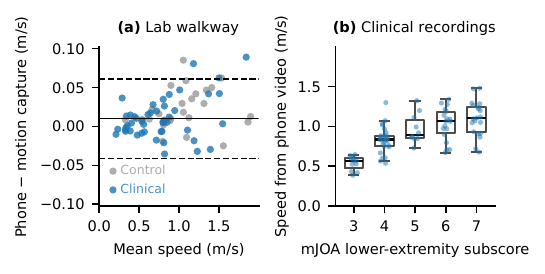}
\vspace{-23pt}
\caption{Walking speed from smartphone video, with the recorded height. (a) Bland--Altman plot against motion capture on the lab walkway, one point per walk; solid line: the clinical walks' bias, dashed: their 95\% limits of agreement. (b) Steady walking speed of patients with cervical myelopathy in the clinic hallway by the same-day mJOA lower-extremity subscore (boxes: quartiles; points: sessions).}
\label{fig-clinical-speed}
\end{figure}

Our calibration-free workflow, which combines foundation models for scene geometry and camera-relative biomechanical
pose with a lightweight sequence solver,
measures walking speed from a handheld smartphone with a median error of 0.016 m/s on held-out clinical walks, and 0.012 m/s with
the recorded height. Both are well below the MCIDs for stroke (0.16 m/s) and incomplete spinal cord injury (0.15 m/s)
\parencite{tilson_meaningful_2010, forrest_are_2014}, and ours is the only method whose error is below even the smallest
change considered meaningful in older adults and people after stroke (0.04 m/s) \parencite{perera_meaningful_2006}. Our method's error is smallest for the slowest
walkers (Figure~\ref{fig-clinical-speed}a), for whom an MCID matters most. Among existing approaches, the learned world-frame methods come closest (FlowHMR 0.051, GVHMR
0.057 m/s), which build the walker's path by adding up per-frame
velocities predicted from the body's pose, leading to drift which we observed in this work (Table~\ref{tab-phone-gait}). The scene-based methods did worse: Human3R, which regresses the camera, scene and people from one network, had
the largest walking-speed error (0.085 m/s), mostly from scale (17.2\%). JOSH also optimizes the person and the camera, but it adds a dense scene, lets the body's
shape change every frame \parencite{liu_josh_2026} and takes five times our runtime (1.6 against 0.3 s per frame). It
kept the shape of the path best of the comparison methods
once scale is removed (RTE 0.69\%, against our 0.36\%), but its scale was off by 8.3\%, so its walking-speed error (0.084 m/s) was nearly as large, and its
feet slid 80 mm.

Our solver outperforms naively scaling the trajectory by a constant or low-pass filtering it
(Table~\ref{tab-phone-variants}). Each lowered
walking-speed error, but even combined they trail the solver in scale, drift and walking speed. With the low-pass filter, the pelvis trajectory is also smoother than the ground truth's (jitter ratio
0.5--0.6), as are GVHMR's and FlowHMR's \parencite{shen_gvhmr_2024, wang_flowhmr_2026} (0.6 and 0.5), which may indicate
oversmoothing. The solver's (0.7) is the closest to the ground truth's.

Earlier approaches keep the camera stationary, regress speed from 2D keypoints (body joint positions detected in
the image) or use wearables. In mean absolute error, a static camera comes within 0.03--0.04 m/s of motion capture from
the side but 0.10--0.15 m/s from the front \parencite{stenum_clinical_2024}, speed regressed from a following camera's
keypoint trajectories within 0.13--0.14 m/s in children with cerebral palsy and prosthesis users
\parencite{kidzinski_deep_2020, cimorelli_prosthesis_2024}, and a lower-back wearable within 0.06--0.12 m/s
\parencite{kirk_mobilise_2024}. Ours is 0.026 m/s on the clinical walks (0.020 m/s with the recorded height). As a mean
signed error $\pm$ SD, keypoint regression after stroke gives $-$0.02 $\pm$ 0.11 m/s \parencite{lonini_video_2022},
against our +0.01 $\pm$ 0.03 m/s with the recorded height, and a pocketed phone has even wider limits of agreement
\parencite{werner_apple_2023}. We reach these errors from a moving handheld phone without calibration or setup, though
none of these comparisons is head-to-head.

Beyond walking speed, our footfalls and pelvis path compare well with smartphone biomechanics systems that need
fixed or calibrated cameras. Two calibrated phones on tripods running OpenCap measured healthy adults' stride length with a bias of about
$-$1 cm against motion capture \parencite{de_borba_walking_2025}. Ours is +1.2 cm on control walks and +0.2 cm on
clinical walks (Table~\ref{tab-spatial-gait}), from one moving handheld phone.
OpenCap Monocular \parencite{gilon_opencap_2026} reports a pelvis translation mean absolute error of 3.4 cm per axis from
a phone on a tripod, in 10 healthy adults performing tasks with less displacement and short level walks within a
laboratory capture volume; ours is 2.1 cm per axis on control walks and
2.4 cm on clinical walks, aligned over the whole walk with no scale. Our footfall and pelvis errors hold on the slow, asymmetric, often device-assisted clinical gait the solver was
never tuned on. OpenCap Monocular was
originally validated only in healthy adults, and in our earlier comparison it degraded the most of the methods compared
on clinical participants \parencite{cotton2026biomechanical}. In terms of generalization, our workflow carried over, with no walkway, setup or
calibration, to two clinic hallways with different lighting, room geometry and texture. There, steady walking speed
graded with the same-day lower-extremity mJOA ($\rho$ = 0.66), more closely than any gait metric from our earlier smartphone pipeline in this cohort
\parencite{peiffer_portable_2026}.

Our workflow could reduce the small overhead of timing a 10-meter walk test, a common clinical measure, while also
providing kinematics that our earlier smartphone pipeline showed to be responsive to clinical intervention
\parencite{peiffer_portable_2026}. Processing takes about 8--9 times the walk's duration on one GPU (under a minute for a
7 s walk), so results could be ready within a visit, and the faster comparison methods trade that time for three to
five times the walking-speed error. The workflow's ease of use could also enable routine monitoring at home, which matters because a
timed clinic test measures capacity rather than daily performance. Older adults walk faster in a clinic test than they
usually do in daily life, with only a weak correlation between the two \parencite{vanancum_gait_2019, takayanagi_relationship_2019},
and after stroke, clinical speed may not distinguish levels of community walking \parencite{bansal_speed_2024}. Here, unstructured walks in
a clinic hallway already tracked the same-day mJOA.

\textbf{Limitations.} First, the solver assumes one flat floor and regular foot contacts, so ramps, stairs and poorly
reconstructed floors may bias the scale. Footfall-based parameters also hinge on a gait-event detector that may not be robust to slow, shuffling
gait. Second, the ground truth
for pelvis velocity is at 29 Hz, and its own noise (Figure~\ref{fig-person-trajectory}) limits the jitter
comparisons, though not walking speed over a walk. Third, our approach estimates usable lens intrinsics with VGGT-$\Omega$, but was tested with only two phone models,
and every video was recorded with the phone on a DJI handheld gimbal, which smooths its motion, so generalization to
unstabilized handheld video remains to be tested. Finally, our joint accuracy is tied to B3D's: the solver changes
its joint angles and positions by under a degree and a millimeter, and it optimizes only B3D's biomechanical skeleton,
whose MHR mesh enters only through the initial metric scale. As in B3D's earlier comparison
\parencite{cotton2026biomechanical}, PBL's joint angles and positions remain more accurate, though PBL recovers no metric
world trajectory, and so no walking speed or step length. Adding detected 2D keypoints to the solver's energy, and
comparing an implicit parameterization of the motion like PBL's \parencite{peiffer_portable_2026} with our explicit one,
are future work. The planted-foot term also did not lower walking-speed error on the
clinical walks (Table~\ref{tab-phone-variants}), so the solver may do better without it, but that choice needs testing
in other cohorts, devices and environments rather than resting on these held-out walks.

\section{Conclusion}

From a handheld smartphone that follows the walker, our workflow measures walking speed in people with neurologic
conditions and lower-limb prostheses to about a hundredth of a meter per second against motion capture, several times
closer than current world-frame human-motion methods, and step length, stride length and step width with biases within about
1 cm. Within routine visits to two neurosurgery clinics, with no walkway or setup, its walking speed tracked the
same-day clinical scores of patients with cervical myelopathy (lower-extremity mJOA, $\rho$ = 0.66). This makes walking speed measurable wherever
a patient can be filmed, enabling routine, real-world tracking of
movement through aging, disease and rehabilitation.

\FloatBarrier
{\small
\printbibliography
}

\end{document}